\documentclass[journal,twocolumn,10pt]{IEEEtran}
\usepackage{amsmath,epsfig} 

\usepackage{tabularx}  
\usepackage{cite}

\usepackage{floatflt} 
\usepackage{paralist} 
\usepackage{algorithm} 
\usepackage{graphicx}
\usepackage{amsthm}
\usepackage{amssymb}
\usepackage{listings}    
\usepackage{xcolor}
\usepackage{hyperref}
\usepackage{indentfirst}
\usepackage{booktabs}
\usepackage{subfigure}
\usepackage{caption}
\usepackage{calligra}
\usepackage{bm}
\usepackage{fancyhdr}
\usepackage{float}
\usepackage{pifont}
\usepackage{colortbl} 
\usepackage{amsmath}
\usepackage{MnSymbol}%
\usepackage{amsbsy}
\usepackage{ulem} 
\usepackage{multicol}
\usepackage{multirow}
\usepackage{mathrsfs}

\newcommand{\etal}{\textit{et al}.}

\def\x{\mathbf{x}}

\def\f{\mathbf{f}}

\def\p{\mathbf{p}}

\def\vv{\mathbf{v}}

\newcommand{\mypar}[1]{{\bf #1.}}
\theoremstyle{definition}

\DeclareMathOperator{\Adj}{A}
\DeclareMathOperator{\Cm}{C}

\DeclareMathOperator{\F}{F}

\DeclareMathOperator{\Mm}{M}

\DeclareMathOperator{\Ss}{S}
\DeclareMathOperator{\Vm}{V}
\DeclareMathOperator{\X}{X}

\DeclareMathOperator{\W}{W}

\newcommand{\R}{\ensuremath{\mathbb{R}}}

\DeclareMathOperator{\Id}{I}

\title{A 3D Mesh-based Lifting-and-Projection Network for Human Pose Transfer}

\author{
Jinxiang Liu*,
Yangheng Zhao*,
Siheng Chen
and Ya Zhang 
\thanks{Jinxiang Liu, Yangheng Zhao, Siheng Chen and Ya Zhang are with the Cooperative Medianet Innovation Center, Shanghai Jiao Tong University, Shanghai 200240, China (e-mail: \{jinxliu, zhaoyangheng-sjtu, sihengc, ya\_zhang\}@sjtu.edu.cn). Ya Zhang is the corresponding author. }

\thanks{* equal contribution.}}

\begin{document}

\maketitle

\begin{abstract}
Human pose transfer has typically been modeled as a 2D image-to-image translation problem. This formulation ignores the human body shape prior in 3D space and inevitably causes implausible artifacts, especially when facing occlusion.  To address this issue, we propose a \emph{lifting-and-projection} framework to perform pose transfer in the 3D mesh space. The core of our framework is a foreground generation module, that consists of two novel networks: a lifting-and-projection network (LPNet) and an appearance detail compensating network (ADCNet).  To leverage the human body shape prior, LPNet exploits the topological information of the body mesh to learn an expressive visual representation for the target person in the 3D mesh space. To preserve texture details, ADCNet is further introduced to enhance the feature produced by LPNet with the source foreground image. Such design of the foreground generation module enables the model to better handle difficult cases such as those with occlusions. 
Experiments on the iPER and Fashion datasets empirically demonstrate that the proposed lifting-and-projection framework is effective and outperforms the existing image-to-image-based and mesh-based methods on human pose transfer task in both self-transfer and cross-transfer settings.
\end{abstract}


\section{Introduction}
\label{sec:intro}

Human pose transfer, also termed pose-guided image generation~\cite{ma2017pose}, has drawn an increasing attention in the computer vision community. Given a source image of a person and a target pose image, this task aims to synthesize a realistic-looking image where the source person executes the target pose with the source visual appearance preserved.
This task has potential applications in a wide range of areas, such as data augmentation \cite{sun2019unlabeled}, virtual try-on \cite{dong2019fw} and video-making \cite{yang2018pose}.


\begin{figure}[tb]
\begin{center}
\includegraphics[width=1.0\linewidth]{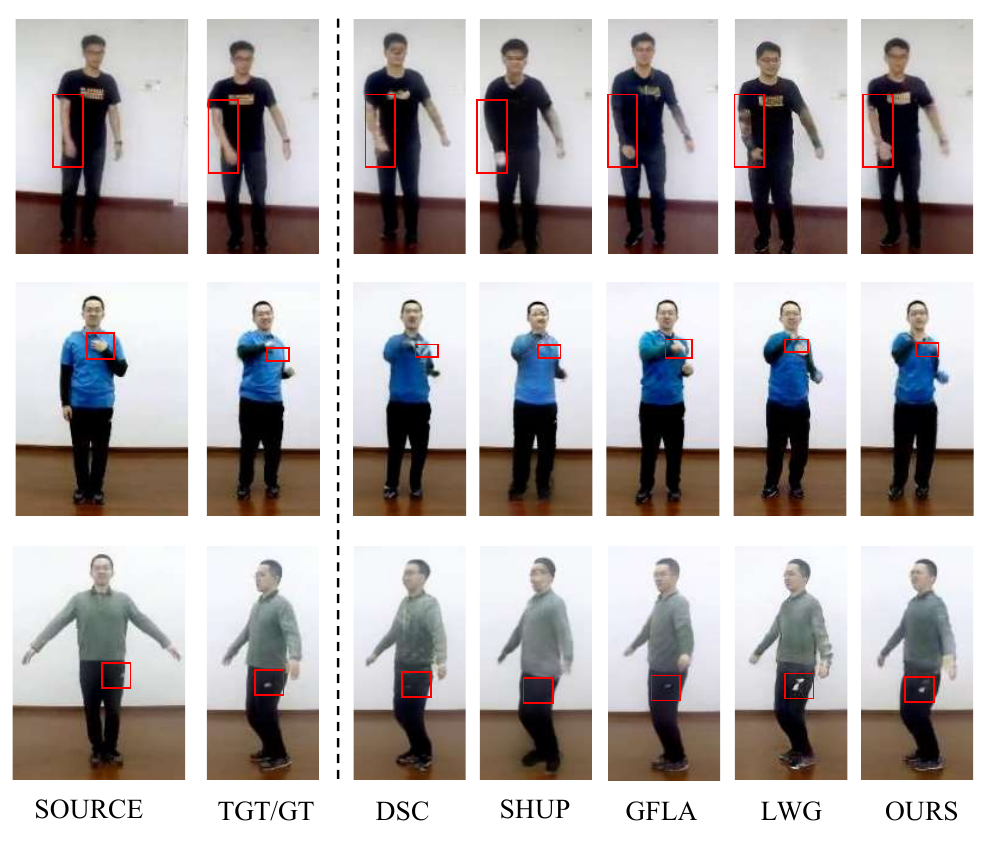}
\end{center}
\caption{Images generated by different methods on the \textit{iPER} dataset.
The ﬁrst column contains source images, and the second column contains ground truth images with target poses. Target images and synthesized images are center-cropped for better presentation. Our method outperforms other approaches in terms of realness and details, especially for some difficult cases such as occlusion, as marked with a red box. 
Our model produces the most plausible results which are closest to the ground-truths.  
}
\label{fig:intro-results}
\end{figure}

The success of human pose transfer critically depends on a proper pose representation. 
As a seminal work of human pose transfer, PG2~\cite{ma2017pose} proposes representing the 2D target pose into its corresponding keypoint heatmap, and it is later followed by many additional studies~\cite{siarohin2018deformable, zhu2019progressive, zhang2020cross}.
However, the sparsity of the keypoint heatmap representation is often found to be insufficient to outline the human body and leads to observable artifacts.
To exploit more spatial layout guidance information from a target pose,
studies~\cite{dong2018soft,song2019unsupervised,liu2019swapgan} introduce an extra human body parsing map~\cite{gong2017look},
which indicates the body semantic segmentation label of each pixel, to guide the generation process. However, a parsing map only provides part-level 2D layout guidance, which is still too coarse.

Benefiting from the advancement of human body modeling, denser pose representations, such as DensePose~\cite{alp2018densepose} and 3D body mesh~\cite{loper2015smpl}, have been explored in the human pose transfer task.
DensePose defines a dense mapping between an image and UV space, which enables pose transfer through an image $\rightarrow$ UV space $\rightarrow$ image mapping process ~\cite{neverova2018dense,grigorev2019coordinate}.
The body mesh~\cite{loper2015smpl}, providing a comprehensive representation for both human shape and pose, has recently been introduced to pose transfer
~\cite{guan2019human,zanfir2018human,sun2020human,liu2019liquid,li2019dense}.
Sun \etal{}~\cite{sun2020human} propose using the projected label map of the reconstructed body mesh as an auxiliary input for the generator. Li \etal{}~\cite{li2019dense} render an appearance flow map and a visibility map to warp the source image. Liquid warping GAN~\cite{liu2019liquid} computes the transformation flow from the source and the target meshes, and warps the feature of the source image to act as the input of the target image generator. 
The aforementioned methods practically leverage the meshes to produce additional guidance information in the 2D plane and fail to fully leverage other essential characteristics that the body mesh carries such as the 3D spatial structure.

This paper aims to explicitly leverage the 3D topological structure of the body mesh for pose transfer. Instead of formulating it as a direct image-to-image translation task, we leverage a 3D body mesh as an intermediary and perform pose transfer in 3D space. 
For this purpose, we introduce a lifting-and-projection network (LPNet), which first maps the human appearance features extracted from the source image to a predefined 3D human mesh model (\emph{i.e.}, lifting), then performs the pose transfer and visual feature learning in 3D space, and ﬁnally projects the features back to a 2D plane for target image generation (\emph{i.e.}, projection).
To facilitate visual feature learning in 3D space, 
LPNet leverages the natural graph structure of the body mesh and adopts graph convolutions ~\cite{kipf2016semi} to propagate and update the visual appearance features for the target pose image along the mesh graph, which leads to a more expressive and contextual-aware feature representation for the pose transferred mesh vertices.
With the assistance of the lifting-and-projection procedure, performing pose transfer and representation learning in the 3D mesh space endows the model with a stronger ability in handling some difficult cases in pose transfer task, such as occlusion problems as shown in Fig. \ref{fig:intro-results}.
To compensate high-frequency detail loss and inconsistency due to the rasterization and the sampling operations in LPNet, inspired by adaptive instance normalization~\cite{huang2017arbitrary}, we further devise an additional appearance detail compensating network (ADCNet), which modulates the output features of LPNet with the source image so that the details of the source image are preserved properly in the synthesized image.

For realistic tasks, we need to handle not only human pose transfer (\emph{i.e.}, foreground generation), but also the preservation of the surrounding background of the person (\emph{i.e.}, background generation). The two sub-tasks are completed using separate modules.
We also introduce a fusion module to merge the synthesized foreground and background images. The entire lifting-and-projection framework for human pose transfer thus includes four modules, namely, a mesh recovery module for pose representation, background generation module, foreground generation module, and fusion module, as shown in Fig. \ref{fig:modules}. 

To validate the effectiveness of our proposed method, we perform extensive experiments on the iPER and Fashion datasets under two transfer settings: self-transfer and cross-transfer. 
The experimental results demonstrate that by incorporating 3D human structural information and processing with graph convolution networks, 
the proposed lifting-and-projection framework outperforms existing human pose transfer methods both visually and quantitatively. 

The main contributions of this paper include the following:
\begin{itemize}
    \item We propose a lifting-and-projection framework to perform human pose transfer in 3D space leveraging 3D human shape prior in the form of 3D mesh, which is expected to better handle difficult cases such as occlusion.
    \item Leveraging the natural graph structure of a 3D body mesh, LPNet adopts graph convolution for representation learning along with the body mesh, which leads to a more expressive and contextual-aware feature representation for the pose-transferred mesh vertices.
    \item We design ADCNet to preserve the detailed texture features in the source image and enhance the synthesized target image quality.
    \item The experimental results demonstrate the effectiveness and superiority of our proposed framework compared with existing methods both quantitatively and qualitatively.
\end{itemize}

\section{Related Works} \label{sec:related}
Human pose transfer has become an increasingly important research direction with many applications including image and video editing, virtual reality and games, etc. 
In this section, we give a detailed review of previous works on human pose transfer.
In addition, we briefly introduce some literature on graph convolutional networks that we used in our method.  

\subsection{Human Pose Transfer}

Generative adversarial networks (GANs)~\cite{goodfellow2014generative}, especially conditional GANs (cGANs) \cite{mirza2014conditional}, have shown superior ability in generating realistic images with high image quality.
The seminal work PG2 \cite{ma2017pose} first represents a given target pose with the corresponding keypoint heatmap and concatenates the heatmap image with the source image as the inputs of an autoencoder network.
The authors also designate a two-stage conditional GAN training strategy to train the whole model.
Many works \cite{balakrishnan2018synthesizing,siarohin2018deformable,zhang2020cross,yang2020towards,yang2020region,albahar2019guided,tang2020xinggan,tang2020bipartite} have used this method to represent a 2D keypoint pose with the corresponding heatmap.
Researchers solve the problem by modeling it as an image-to-image translation task and are devoted to designing delicate networks for 2D processing.
Other works \cite{siarohin2018deformable,balakrishnan2018synthesizing} aim to learn the spatial transformations of each body part at the feature level based on pose differences.
PATN \cite{zhu2019progressive} and its variants \cite{yang2020region,yang2020towards,li2020pona}  employ cascaded blocks to generate target images progressively, and the image branch and pose branch within each generation block interact with each other to learn a better presentation.
By treating human pose transfer as a 2D image-to-image translation problem, \cite{albahar2019guided} and \cite{zhang2020cross} focus on the similarities among different image-to-image tasks, and establish unified frameworks to tackle these tasks.

Although some success has been achieved in human pose transfer by viewing it as a 2D image-to-image translation problem, how to leverage more information on the human body to obtain more realistic results remains to be explored, such as the use of human parsing maps \cite{liu2015fashion,gong2017look}, DensePose \cite{alp2018densepose,neverova2018dense} and body meshes \cite{loper2015smpl}.

A human parsing map  \cite{liu2015fashion,gong2017look,xia2017joint} provides a hint of the human body semantic category of each pixel in an image; hence, researchers \cite{song2019unsupervised,liu2019swapgan} introduce it as spatial layout guidance of humans in human pose transfer task.
Generally, the researchers first deploy a network to predict the human parsing map for the source image or the target pose image and use the predicted human parsing map as an extra input besides the keypoint heatmap to synthesize the target image.

In addition to using 2D representations of human poses and shapes including keypoint heatmaps and human parsing maps, a group of researchers intend to incorporate advanced 3D human modeling methods, 
such as DensePose and human body mesh, to guide the pose transfer process. 
Neverova \etal{} \cite{neverova2018dense} send a concatenation of the source dense pose, target dense pose and source image to an autoencoder network to predict a coarse image, and then deploy another autoencoder network to inpaint the full texture of the human body.  
Different from \cite{neverova2018dense} which directly inpaints texture with an auto-encoder, Grigorev \etal{} \cite{grigorev2019coordinate} build a network model to estimate the source location of each pixel in a human model UV map, thereby establishing the correspondence between the source and target image. 

In contrast to DensePose-based methods which directly bridge the position of the human part 2D image plane and 3D UV space, mesh-based methods  \cite{zanfir2018human,guan2019human,li2019dense,sun2020human,liu2019liquid} typically undergo mesh reconstruction.
However, the use of the reconstructed human body mesh of these methods for human pose transfer varies greatly.
Zanfir \etal{}  \cite{zanfir2018human} and Guan \etal{} \cite{guan2019human} use barycentric transformation to render the appearance of the target image. Sun \etal{} \cite{sun2020human} use a reconstructed body mesh to render a label map as a condition for the conditional GAN setting. 
Liquid warping GAN \cite{liu2019liquid} renders a correspondence map and a transformation map to sample and warp the source image into target image. 
However, the drawback of these methods is that few of them explicitly exploit the integral 3D structure information of the reconstructed body mesh which is crucial in human modeling.

The core difference between our method and the existing mesh-based methods is that we utilize the 3D structure of the body mesh to promote human pose transfer. The existing mesh-based methods leverage meshes to produce label maps\cite{sun2020human}, correspondence maps and warping maps\cite{liu2019liquid}, which are all pose transfer guidance represented in 2D maps.
In contrast, the proposed method aims to exploit the 3D structure of the body mesh and perform transfer and feature learning along the 3D meshes.
Concretely, we lift the visual features onto a 3D human mesh through sampling, and employ graph convolutional networks to learn target visual features in the 3D mesh space,  
which boosts the performance of human pose transfer.

\subsection{Graph Convolutional Networks for Mesh Modeling}
Convolutional neural networks (CNNs) have shown its powerful representation ability in images and are ubiquitous in many vision tasks such as classification, detection and segmentation \cite{krizhevsky2017imagenet,girshick2014rich,long2015fully}. 
However, due to the regularity property, CNNs are less capable of dealing with nonstructured data such as mesh and point clouds. 
To address these problems, researchers have turned to graph convolutional networks (GCNs) \cite{kipf2016semi} for nonstructured data, and many works have investigated the mesh modeling with GCNs. 
Litany \etal{} \cite{litany2018deformable} design a variational autoencoder with graph convolutional operations aiming to learn latent space for complete realistic shapes and then use it for mesh reconstruction from partial scans.
In \cite{ranjan2018generating}, the authors propose learning a nonlinear representation of the face mesh by means of graph spectral convolutions on a face mesh surface. 
A major advance is GraphCMR \cite{kolotouros2019convolutional}, which targets 3D body mesh reconstruction from single monocular 2D image. 
The authors encode the input image into a low-dimension vector, and then, a mesh graph is established where each vertex feature is the concatenation of template vertex coordinates and the image embedding. 
Finally, through graph convolutions, the coordinates of each vertex are regressed with weak supervision.
Additionally, some works \cite{ge20193d} \cite{kulon2019single} utilizes graph convolutions to model hand mesh considering the similarities between the hand mesh and body mesh.  
All the above works basically model either a human body mesh or face mesh as a graph and adopt graph convolutions to learn more accurate shapes or structural information. 

Different from the above studies that deploy graph convolutions to reconstruct a more accurate and plausible mesh, our work intends to learn a contextual visual representation of the target person by accounting the structural information of the body mesh with graph convolutions. 
To the best of our knowledge, this is the first work to address the human pose transfer problem with graph convolution networks over 3D meshes.

\section{Methodology}
\label{sec:networks}

\begin{figure*}
\begin{center}

  \includegraphics[width=0.9\linewidth]{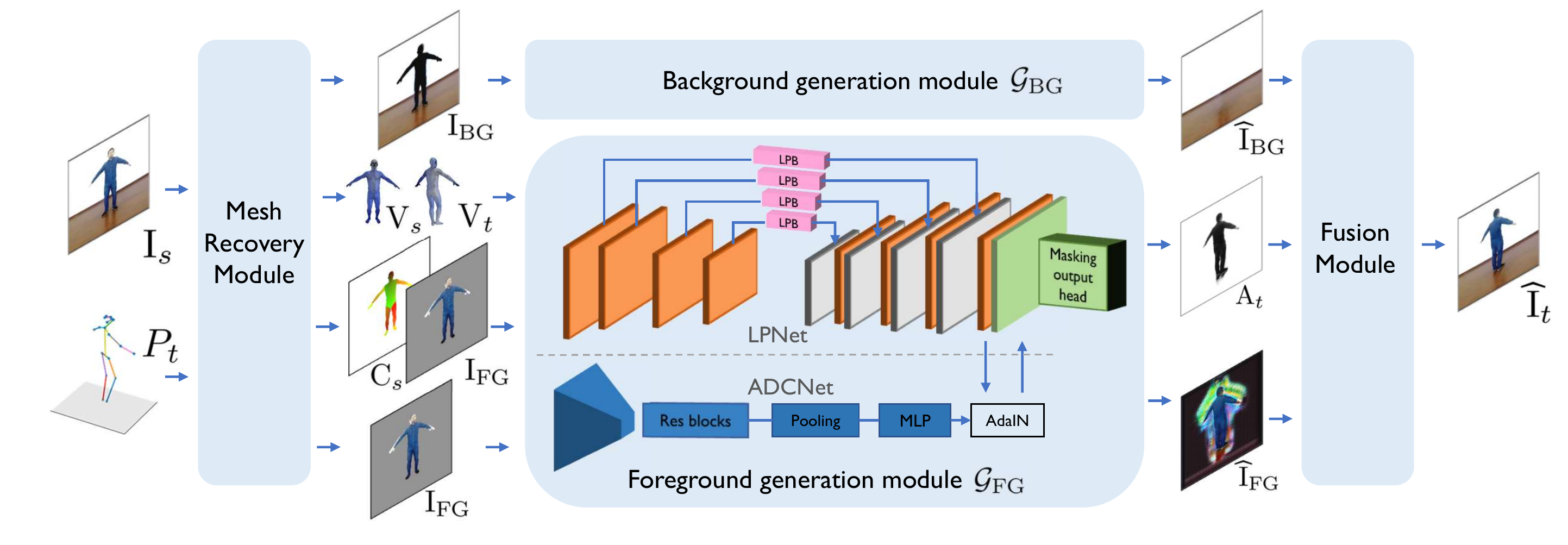}
\end{center}
\caption{ Overall pipeline of our proposed lifting-and-projection framework, which consists of four modules: a mesh recovery module for source and target person mesh reconstruction, background generator $\mathcal{G}_{\rm BG}$ for background inpainting, foreground generator $\mathcal{G}_{\rm FG}$ for foreground generation and fusion module, which merges the results of $\mathcal{G}_{\rm BG}$ and $\mathcal{G}_{\rm FG}$ to produce the final result.
}  
\label{fig:modules}
\end{figure*}

In this section, we present a novel lifting-and-projection framework for human pose transfer.
Section~\ref{sec:overview} first formulates the problem and gives a brief overview of our proposed model.
Then, Sections~\ref{sec:mrm},~\ref{sec:bg},~\ref{sec:fg}, and~\ref{sec:fusion} present the mesh recovery module, background generation module, foreground generation module and background-foreground fusion module, respectively.
Finally, Section~\ref{sec:training} introduces the loss functions for model training.

\subsection{Problem formulation and model overview}
\label{sec:overview}
Given a source image $\Id_s$ and a target image $P_t$, the goal of human pose transfer is to synthesize an image $\widehat{\Id}_t$ that preserves the identity and appearance of the person in source image $\Id_s$ and conforms to the given pose in the target image $P_t$ at the same time.
Here, the target pose $P_t$ contains 1) the rotation angle of the body joint in the target image, $\theta_t \in \mathbb{R}^{24 \times 3}$; and 2) the camera parameter $K_t \in \mathbb{R}^3$, which reflects the translation and scale of a human body in the target image.

To solve the task of human pose transfer, our core ideas are as follows: i) We split the background and foreground generation. The intuition behind this idea is that since the deformation of a human shape mostly changes the foreground and leaves the background untouched, it is more effective to handle the foreground and background generations separately; and ii) 
We leverage the structural information of the reconstructed mesh to promote the learning of foreground feature deformation in 3D space.
Since the 3D mesh is a holistic human representation, feature learning in 3D space can tackle some difficult cases that are challenging to solve in a 2D plane, such as occlusion.

To achieve these goals, our system includes four modules, namely, a mesh recovery module, background generation module, foreground generation module and background-foreground fusion module; see an illustration of the overall architecture in Fig. \ref{fig:modules}. The mesh recovery module reconstructs 3D meshes in both source and target poses for the subsequent modules from an input image and a target pose; the background generation module focuses on generating the complete background image; and the foreground generation module focuses on generating the area around a human body based on 3D human meshes. The background-foreground fusion module produces the final output by fusing the outputs from the background generation module and the foreground generation module.

We elaborate on each of these four modules in the following sections.

\subsection{Mesh Recovery Module (MRM)}
\label{sec:mrm}

In the human pose transfer task, the original inputs are 2D RGB images. To leverage a 3D mesh, we need to recover 3D meshes from the given images first.
Given the source image and the target pose as inputs, the mesh recovery module is responsible for reconstructing both the source and target meshes.
In this module, we harness the pretrained HMR~\cite{kanazawa2018end} model, which is an end-to-end neural-network-based system, to recover a 3D full body mesh from a single monocular RGB image.
HMR achieves 3D mesh recovery based on two steps. First, for a given image, HMR estimates the shape parameters $\beta \in \mathbb{R}^{10}$, reflecting the body shape characteristics of the detected person, the pose parameters $\theta \in \mathbb{R}^{24\times 3}$, reflecting the relative 3D rotation of body joints in a certain pose, and the camera view $K$, reflecting the translation and scale of the 3D mesh.
Second, HMR recovers the 3D mesh and obtains the vertex coordinates through the skinned multi-person linear model (SMPL)~\cite{loper2015smpl}, which is a triangular human mesh model that can utilize the shape parameters and the pose parameters to construct a human body surface with $N=6890$ vertices.

\textbf{Source Mesh Recovery.} Given a source image $\Id_s$, it is straightforward to recover the source body mesh $\Vm_s$ based on HMR. HMR takes $\Id_s$ as the input and estimates camera view parameters $K_s$ as well as the shape parameters $\beta_s$ and pose parameters $\theta_s$ for the detected person; that is,
\begin{equation}
    \beta_s, \theta_s, K_s = \operatorname{HMR}\left(\Id_s \right).
\end{equation}
Then, the 3D body mesh of source person $\Vm_s$ can be readily reconstructed with the differentiable function of the SMPL model as
\begin{equation}\label{eq:smesh}
   \Vm_s = m(\theta_s, \beta_s) \in \mathbb{R}^{N \times 3},
\end{equation}
where $m(\cdot,\cdot)$ is the linear blend-skinning function of SMPL.

\textbf{Target Mesh Recovery.} To reconstruct the target body mesh $\Vm_t$, it is required to combine both the shape parameters estimated from source image $\Id_s$ and the given target pose parameters $\theta_t$. The shape parameters $\beta_s$ estimated from the source image $\Id_s$ preserve the human shape information for the same person, while the target pose $P_t$ contains the target pose parameters $\theta_t$ and the camera view parameters $K_t$. Then, the target body mesh $\Vm_t$ is constructed by composing $\beta_s$ and $\theta_t$ via the SMPL model; that is,
\begin{equation}
\label{eq:tmesh}
    \Vm_t = m(\theta_t, \beta_s) \in \mathbb{R}^{N \times 3},
\end{equation}
where $m(\cdot,\cdot)$ is the same linear blend-skinning function as Eq.~\ref{eq:smesh}.

Finally, the mesh recovery module outputs the source pair of the 3D mesh and the camera view ($\Vm_s$,$K_s$) and the target pair ($\Vm_t$,$K_t$), which will be consumed in the following modules.

\subsection{Background Generation Module }
\label{sec:bg}
Since the deformation of a human pose almost changes the foreground of an image and leaves the background untouched, we will handle the foreground and background generations separately.
The overall image generator $\mathcal{G}$ thus includes two parts:
the background generator $\mathcal{G}_{\rm BG}$ and foreground generator $\mathcal{G}_{\rm BG}$. Following \cite{liu2019liquid}, we separate the input source image $\Id_s$ into the foreground $\Id_{\rm FG}$ and the background $\Id_{\rm BG}$ based on the silhouette of the source mesh $\Vm_s$.

To generate the background, we employ an image inpainting network as $\mathcal{G}_{\rm BG}$, which aims to hallucinate the background areas occluded by the human body in the source image. The input of $\mathcal{G}_{\rm BG}$ is the body-masked source image $\Id_{\rm BG}$, and the output is a background-inpainted image $\widehat{\Id}_{\rm BG}$. This image inpainting network is implemented with a vanilla CNN encoder-decoder network consisting of three stride-2 convolution layers for downsampling, six residual blocks~\cite{he2016deep} and three stride-2 transconvolution layers for upsampling.

\subsection{Foreground Generation Module }
\label{sec:fg}

\begin{figure*}
\begin{center}
  \includegraphics[width=1\linewidth]{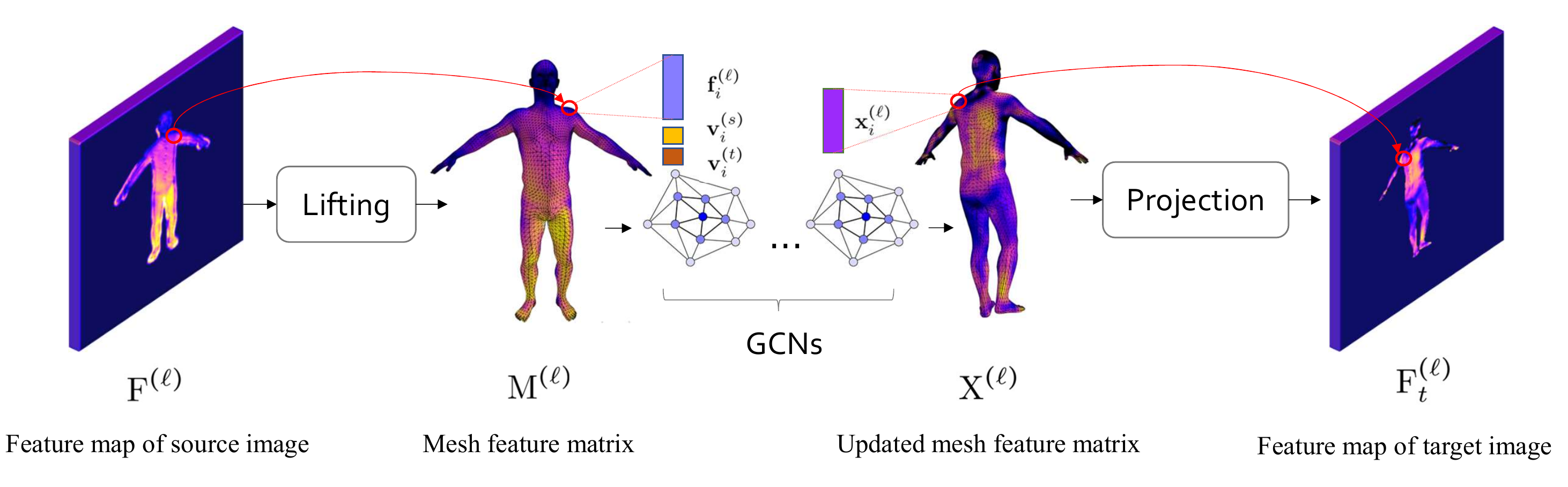}
\end{center}
\caption{ Illustration of lifting-and-projection block (LPB): basic building block of LPNet.
This block consists of three successive steps: a). The \textbf{lifting} step lifts features from the source image feature map $\F^{(\ell)}$ to mesh $\Mm^{(\ell)}$; b). The \textbf{3D processing} step updates the mesh feature matrix $\Mm^{(\ell)}$ with graph convolutional networks; c). The \textbf{projection} step projects the updated mesh feature $\X^{(\ell)}$ to obtain a feature map of target image $\F^{(\ell)}_t$. Here, $\f_i^{(\ell)}$, $\vv_i^{(s)}$, $\vv_i^{(t)}$ and $\x_i^{(\ell)}$ denote the lifted feature, coordinates in the source mesh, coordinates in the target mesh and updated feature of vertex $i$, respectively.}
\label{fig:lpnet}
\end{figure*}

Since the 3D human mesh is a holistic representation of a human body, it is the ideal space to learn the deformation of appearance.
To leverage a 3D human mesh, we propose a novel foreground generator $\mathcal{G}_{\rm FG}$ that consists of two bone components: a lifting-and-projection network (LPNet) and an appearance detail compensating network (ADCNet), as shown in Fig. \ref{fig:modules}.
Pose transfer and target appearance feature learning in the mesh space are achieved by the three core operations of LPNet: lifting, 3D processing and projection. Additionally, an auxiliary ADCNet is designed to enhance the features produced by LPNet by directly working on the source foreground images.

The following sections are the detailed introduction of our proposed foreground generator.

\subsubsection{Lifting-and-Projection Network}\label{se:lpnet}
\label{sec:lift-proj}

The core function of LPNet is to perform pose transfer and learn target foreground features with the assistance of the reconstructed body mesh.

First, we extract visual features at different scales from the source foreground image.
We feed the network with the concatenation of source foreground image $\Id_{\rm FG} \in \mathbb{R}^{H\times W\times 3}$ and the source coordinate map $\Cm_s \in \mathbb{R}^{H\times W\times 3}$. $\Cm_s$ is produced by rasterizing the coordinates of the source mesh vertices $\Vm_s$ to the 2D plane according to $K_s$, which establishes the correspondence between a pixel and a 3D mesh vertex. The three coordinate values of each pixel in the body area of $\Cm_s$ denote its coordinates in 3D space, while other pixel values are all set to zero. We append the coordinate map as an auxiliary input of the feature extractor for the sake of integrating more structural information into the foreground features.

Then, a cascade of novel lifting-and-projection blocks (LPBs) are employed to process the multiscale feature maps, which implement the three core operations of LPNet: lifting, 3D processing and projection. An illustration of LPB is shown in ~Fig. \ref{fig:lpnet}.
For a given image feature map at a specific scale, LPB lifts the features to the 3D space based on the pixel-to-mesh-vertex correspondence, uses a graph convolutional network to update visual features over 3D meshes, and finally projects the updated features on 3D meshes back to a 2D plane.

The projected feature maps at different scales are fed into a decoder network to obtain a final foreground feature in the target pose. Here, we use three successive stride-2 convolutional layers to extract the feature maps at four different scales, which are processed by four independent LPBs, and symmetrically use three stride-2 transconvolutional layers with residual blocks \cite{he2016deep} as the decoder network.

In the following paragraphs, we discuss three essential operations of LPNet in detail.

\mypar{Lifting}
Here, we aim to lift the human visual feature in the 2D image space to the 3D mesh space. To achieve this goal, we need correspondence mapping from 2D pixels to 3D mesh vertices; that is, based on the source camera parameter $K_s$, we can project the 3D coordinates of each mesh vertex to the source image and obtain its 2D coordinates. Given the 2D coordinates, we perform a bilinear sample to sum the feature vector of the 4 nearest pixels and assign this feature vector to the mesh vertex. In this way, the 3D mesh vertices share the same visual features with the corresponding 2D pixels; in other words, we lift the visual features from the 2D image to the 3D mesh. Mathematically, let $\F^{(\ell)} \in \mathbb{R}^{H_{\ell} \times W_\ell \times C_\ell}$ be the feature map of the source image at the $\ell$th scale with $H_\ell,W_\ell,C_\ell$ being the height, width and number of channels. For the $i$th 3D mesh vertex, $\vv_i \in \R^3$, the corresponding feature vector is obtained as
\begin{eqnarray}
    \p_i & = & \operatorname{Proj}_{3 \rightarrow 2}\left(\vv_i, K_s \right) \in \R^2,
\\
    \f_i^{(\ell)} & = & \operatorname{Sample} \left(\F^{(\ell)}, \p_i \right) \in \R^{C_\ell},
\end{eqnarray}
where $K_s$ is the source camera view parameter, $ \p_i$ is the corresponding 2D pixel, $\operatorname{Proj}_{3 \rightarrow 2}(\cdot)$ is an association operator that projects a 3D mesh vertex to a 2D pixel and $\operatorname{Sample}(\cdot)$ is the bilinear sampling operation. We can collect the feature vector for all the mesh vertices and form a mesh feature matrix $\Mm^{(\ell)} \in \R^{N \times {C_\ell}}$ whose $i$th row is $\f_i^{(\ell)}$ with $\ell$ the image scale.

\mypar{3D Processing with Graph Convolutional Networks}
Here, we propagate the mesh features over the human mesh and make them adapt to the 3D human body. We first establish a graph based on the innate structure of a human mesh, as the human mesh has already predefined a triangular mesh for the human surface. Each node of the graph is a vertex of the human mesh. To construct the graph adjacent matrix $\Adj \in \{0,1\}^{N \times N}$,
the $(i,j)$th element, $\Adj_{ij}=1$ when vertices $i$ and $j$ are in the same mesh triangle and $0$ otherwise. To explicitly leverage the graph topology, we consider graph convolutional networks (GCNs)~\cite{kipf2016semi} to update the mesh features. To employ both the appearance and geometric information of each mesh vertex, we concatenate the relative coordinates of source mesh $\Vm_s$, the relative coordinates of target mesh $\Vm_t$ and the sampled mesh feature matrix $\Mm^{(\ell)}$ to be the input node feature matrix $\X^{(\ell)} = [\Vm_s, \Vm_t, \Mm^{(\ell)}] \in \R^{C_{\ell}+6}$ for the first graph convolution layer. The updated features after each graph convolution layer are
\begin{equation}
    \X^{(\ell)}  \leftarrow {\rm IN} \left( {\rm ReLU}\left(\Adj \X^{(\ell)}  \W\right) \right),
\end{equation}
where $\W \in \mathbb{R}^{C_{\ell} \times C_{\ell}'}$ is a trainable weight matrix, ${\rm ReLU}(\cdot)$ is the ReLU activation function and ${\rm IN}(\cdot)$ is the
instance normalization. We utilize $4$ graph convolution layers for each LPB, and each layer is connected with a residual connection except for the first layer.

The graph convolution layers enable information diffusion over 3D meshes, and the resulting mesh features aggregate the visual information of neighboring vertices, thus providing an enhanced contextual representation of the target person in 3D space. This feature learning in 3D space also resolves the notorious occlusion problem~\cite{li2019dense}, which is shown in Fig.~\ref{fig:intro-results}.

\mypar{Projection}
To synthesize the final image, we project the pose-transferred visual features processed on mesh space back to a 2D image plane.
This can be realized by some differentiable rasterization operations~\cite{loper2014opendr,kato2018neural} in computer graphics.
This rasterization operation guarantees that the network can be trained end-to-end by defining gradients to enable back-propagation in a neural network.
Here, we employ the differentiable rasterizer~\cite{kato2018neural} to rasterize the features of 3D mesh vertices $\X^{(\ell)}$ to the feature maps $\F_t$ in the 2D image plane.
The total projection operation can be formulated as
\begin{equation}
\label{eq:rasterize}
     \F_t^{(\ell)} = \operatorname{Rasterize}\left(\X^{(\ell)}, \Vm_t, K_t\right),
\end{equation}
where $\X^{(\ell)}$ denotes the output mesh feature matrix in the $\ell$ th scale, $V_t$ are the vertices' coordinates of the target mesh, $K_t$ is the target camera view parameters and $\operatorname{Rasterize}(\cdot)$ is the differentiable rasterization operation.

\subsubsection{Appearance Detail Compensating Network}
In Section~\ref{se:lpnet}, we introduced the design of LPNet in a detailed manner.
While LPNet effectively utilizes the structural information of the human body in 3D space, the rasterization and downsampling operations within LPNet may cause some loss of high-frequency signals of the source image. Therefore, unrealness and inconsistency may occur in human face details and clothing texture details of the generated target image.

To mitigate the detail loss and enhance the texture of the final target image, inspired by recent progress \cite{park2019semantic,kim2020transfer,jiang2020psgan,yang2020region,zhu2020sean,kowalski2020config} on the image synthesis of AdaIN \cite{huang2017arbitrary}, we design an auxiliary branch: appearance detail compensating network (ADCNet).
The basic idea of ADCNet is to distill the high-frequency information from the source image and inject it into the target image with the adaptive instance normalization (AdaIN) technique \cite{huang2017arbitrary}.

Concretely, the foreground source image $\Id_{\rm FG}$ $\in \mathbb{R}^{H_0 \times W_0 \times C_0}$ is first passed through one convolutional layer, two ResBlock layers and a global average pooling layer.
Then, the extracted feature vector that encodes the visual detail information of the source image goes through two MLPs to produce $\alpha \in \mathbb{R}^{C} $ and $\gamma \in \mathbb{R}^{ C}$. $\alpha$ and $\gamma$ are the affine parameters to transform target visual features $h \in \mathbb{R}^{H\times W\times C }$.

The process to enhance target feature maps can be formulated as shown in Eq.~\ref{eq:adc}:

\begin{equation}\label{eq:adc}
    h^{new}_{x,y,c}=\gamma\left(\frac{h_{x,y,c}-\mu_{c}}{\sigma_{c}}\right)+\alpha ,
\end{equation}
where $h_{x,y,c}$ are features before normalization, $h^{new}_{x,y,c}$ is its updated features, and $\gamma$ and $\alpha$ are learned scale and shift modulation parameters from masking source image $\Id_{\rm FG} $. $\mu_c$ and $\sigma _c$ denote the mean and standard deviation of $h_{x,y,c}$, respectively, in each channel. More formally:

\begin{equation}
\begin{aligned}
\mu_{c} &=\frac{1}{H W} \sum_{x}^{H} \sum_{y}^{W} h_{x, y, c} ,\\
\sigma_{c}^{2} &=\frac{1}{H W} \sum_{x}^{H} \sum_{y}^{W}\left(h_{x, y, c}-\mu_{c}\right)^{2}+\epsilon ,
\end{aligned}
\end{equation}
where $\epsilon$ denotes the stability parameter, which prevents divide-by-zero in (\ref{eq:adc}). We set $\epsilon =10^{-5}$ in our implementation.

The affine transformed feature maps are sent to the masking output head to produce the output of the foreground.

\subsubsection{Masking Output Head}
In this stage, we employ a CNN to decode these feature maps to generate an RGB image $\widehat{\Id}_{\rm FG} \in \mathbb{R}^{W\times H \times 3}$ as a foreground image as well as a soft mask image $\Adj_t\in [0,1]^{W\times H}$, whose element reflects the likelihood of being a foreground for each pixel.

The output activation functions for $\widehat{\Id}_{\rm FG}$ and $\Adj_t$ are ${\rm tanh}(\cdot)$ and ${\rm sigmoid}(\cdot)$, respectively.

\subsection{Foreground and Background Fusion} \label{sec:fusion}
Combining the background $\widehat{\Id}_{\rm BG}$ synthesized by $G_{\rm BG}$ and foreground $\widehat{\Id}_{\rm FG}$ synthesized by $G_{\rm FG}$ with the mask $\Adj_t$, we obtain the final synthesized target pose transferred image $\widehat{\Id}_t$:
\begin{equation}\label{eq:comb_bg_fg}
\centering
    \widehat{\Id}_t \ = \ \widehat{\Id}_{\rm FG} \odot \Adj_t + \widehat{\Id}_{\rm BG} \odot (1-\Adj_t),
\end{equation}
where $\odot$ is the elementwise multiplication operation.

\begin{figure*}[h]
\begin{center}
\includegraphics[width=1.0\linewidth]{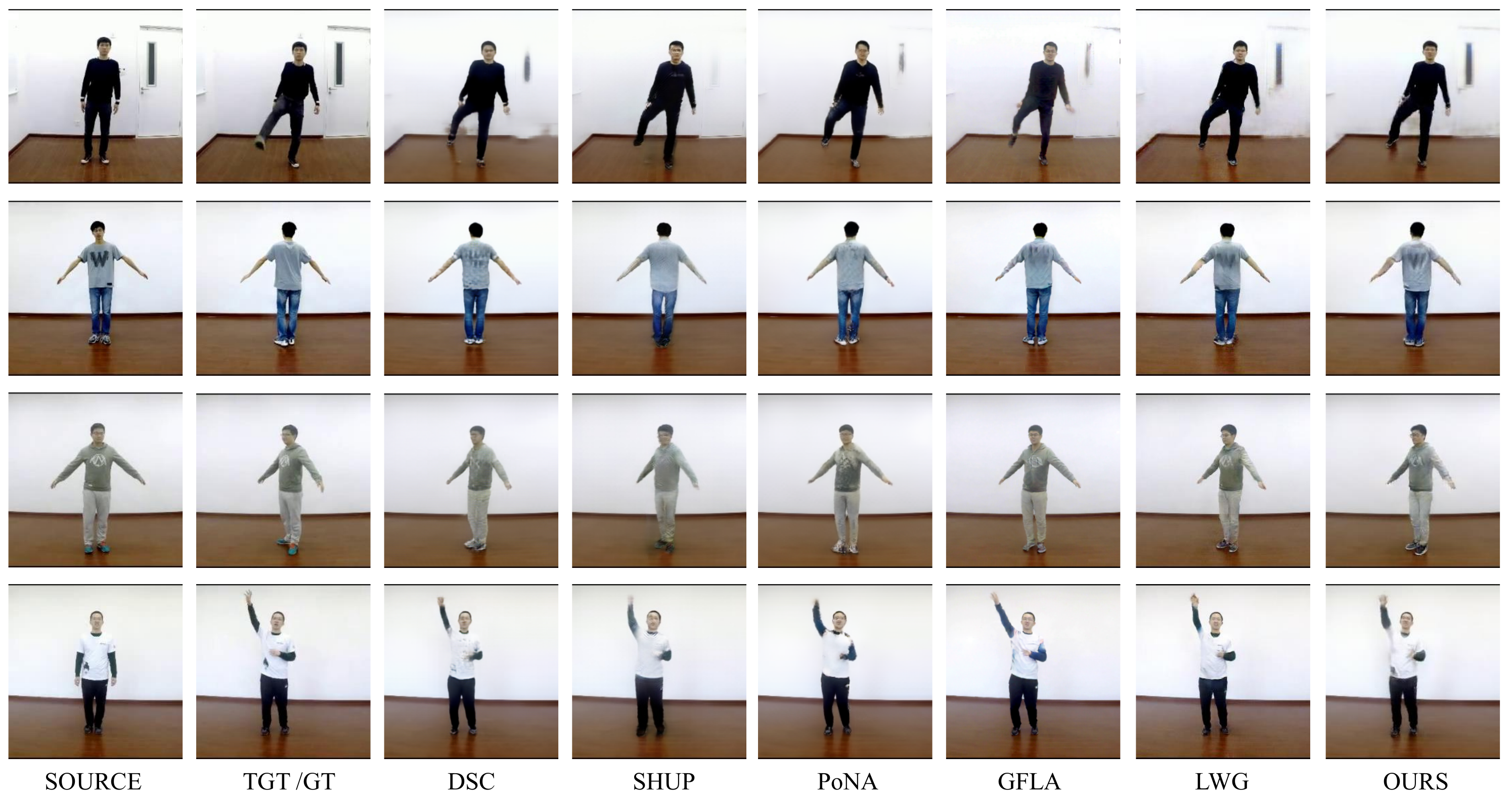}
\end{center}
\caption{Qualitative comparison of our proposed method again existing representative methods on \textit{iPER} dataset in self-transfer setting. Our method preserves more foreground details and is closest to the ground truths.}
\label{fig:vis-si}
\end{figure*}

\subsection{Loss Functions}
\label{sec:training}
In this section, we introduce the loss functions and training strategies for our model.
During the training phase, we randomly sample a pair of images from one training video and set one of them as source image $\Id_s$ and the other as target image $\Id_t$. At each training step, we feed the model with $(\Id_s, P_t)$ and $(\Id_s, P_s)$ to generate target image $\widehat{\Id}_t$ and reconstruct source image $\widehat{\Id}_s$ separately.

We follow the generative adversarial network (GAN) paradigm and employ the PatchGAN discriminator~\cite{pix2pix2016} as our discriminator $\mathcal{D}$. We employ multiple loss functions on $\widehat{\Id}_s$ and $\widehat{\Id}_t$ to train the model simultaneously. The details of the loss functions are discussed below.

\subsubsection{Reconstruction Loss}
To make the training steadier and converge better,
we minimize the $\ell_1$ distance between the reconstructed image $\widehat{\Id}_s$ and real source image $\Id_s$. The reconstruction loss $\mathcal{L}_{\rm rec}$ can be formulated as
\begin{equation}
    \mathcal{L}_{\rm rec} = \left\| \Id_s - \widehat{\Id}_s \right\|_{1}.
\end{equation}

\subsubsection{Perceptual Loss}
The perceptual loss was proposed by Johnson \etal{} \cite{johnson2016perceptual}. It encourages the generated image $\widehat{I}_t$ to have similar feature representation to the ground-truth target image $I_t$ while being not too sensitive to slight spatial misalignments.
We use an ImageNet-pretrained VGG \cite{Simonyan15} network to extract features from the image and minimize the $\ell_1$ distance in the VGG feature space. It is formulated as follows:
\begin{equation}
    \mathcal{L}_{\rm perc} = \sum_{j=1}^{N}\left\|\phi_{j}\left(\widehat{\Id}_{t}\right)-\phi_{j}\left(\Id_{t}\right)\right\|_{1},
\end{equation}
where $\phi_j(\cdot) $ denotes the $j$ th layer's output feature map of the VGG network.

\subsubsection{Adversarial Loss}
The adversarial loss constrains the distribution of generated images to be close to that of the real images~\cite{goodfellow2014generative}.
We adopt the paradigm of conditional GAN \cite{mirza2014conditional}, concatenate the generated image $\widehat{\Id}_t$ and coordinate map $\Cm_t$ as the input of the discriminator, and use the least-square adversarial loss~\cite{lsgan}. Thus, the optimization goal of discriminator is
\begin{align}
    \mathcal{L}^{\mathcal{D}}_{\rm  adv}=\mathbb{E}_{\Id_t \sim p_{data}}(\mathcal{D}(\Id_t, \Cm_t)-1)^{2} 
    +\mathbb{E}_{\widehat{\Id}_t \sim p_{gen}}(\mathcal{D}(\widehat{\Id}_t, \Cm_t)+1)^{2},
\end{align}
while the optimization goal of generator is
\begin{equation}
     \mathcal{L}^{\mathcal{G}}_{\rm  adv}=\mathbb{E}_{\Id_s \sim p_{data}}(\mathcal{D}(\mathcal{G}(\Id_s, P_t), \Cm_t))^{2},
\end{equation}
where $\mathcal{G}$ and $\mathcal{D}$ denote the generator and discriminator, respectively.

\subsubsection{Mask Loss}
Due to the lack of ground-truth annotation for the mask $\Adj_t$, its training relies on the backpropagated gradient from the above losses. However, this situation can lead to a local minimum where the value of $\Adj_t$ is all 1. Such degeneration will block the gradient propagation of $G_{\rm BG}$ and keep it in a suboptimal status. To avoid this issue, following \cite{liu2019liquid}, we use $\Ss_t$, the silhouette of the target mesh obtained by projecting it into the image plane, as a pseudolabel for mask $\Adj_t$ and apply total variation loss to make $\Adj_t$ smoother.

\begin{equation}
    \mathcal{L}_{\rm  mask} = \left\| \Adj_t - \Ss_t\right\|_2^2 + \operatorname{TV}\left( \Adj_t \right),
\end{equation}
where $\operatorname{TV}\left(\cdot \right)$ denotes total variation loss. The formulation is
\begin{equation}
    \operatorname{TV}\left(\Adj \right) = \sum_{i, j}[\Adj_{i, j}- \Adj_{i-1, j}]^{2}+[ \Adj_{i, j}- \Adj_{i, j-1}]^{2}.
\end{equation}

\subsubsection{Face Identity Loss}
To ensure that the generated human images have fewer artifacts on the synthesized face areas,
following \cite{liu2019liquid}, we adopt SphereFaceNet\cite{SphereFaceNet} as a facial feature extractor and apply it to the cropped faces from generated and real images. Then, the $\ell_1$ distance between the generated face image and the real face image is minimized in the face feature space.
\begin{equation}
    \mathcal{L}_{\rm  face} = \left\| g\left(\widehat{\Id}_t\right),g\left(\Id_t\right)\right\|_{1},
\end{equation}
where $g\left( \cdot \right)$ denotes the SphereFaceNet.

The overall loss function for the generator is 
\begin{align}
    \mathcal{L}^{\mathcal{G}} = \, & \lambda_{\rm rec} \mathcal{L}_{\rm rec} + \lambda_{\rm  perc} \mathcal{L}_{\rm perc} + \lambda_{\rm  adv} L^{\mathcal{G}}_{\rm adv} \notag\\
    &+ \lambda_{\rm  mask} \mathcal{L}_{\rm  mask} + \lambda_{\rm  face} \mathcal{L}_{\rm  face},
\end{align}
where $\lambda_{\rm rec}$, $\lambda_{\rm perc}$, $\lambda_{\rm adv}$, $\lambda_{\rm mask}$ and $\lambda_{\rm face}$ are hyperparameters.

\begin{figure*}[ht]
\begin{center}
\includegraphics[width=1.0\linewidth]{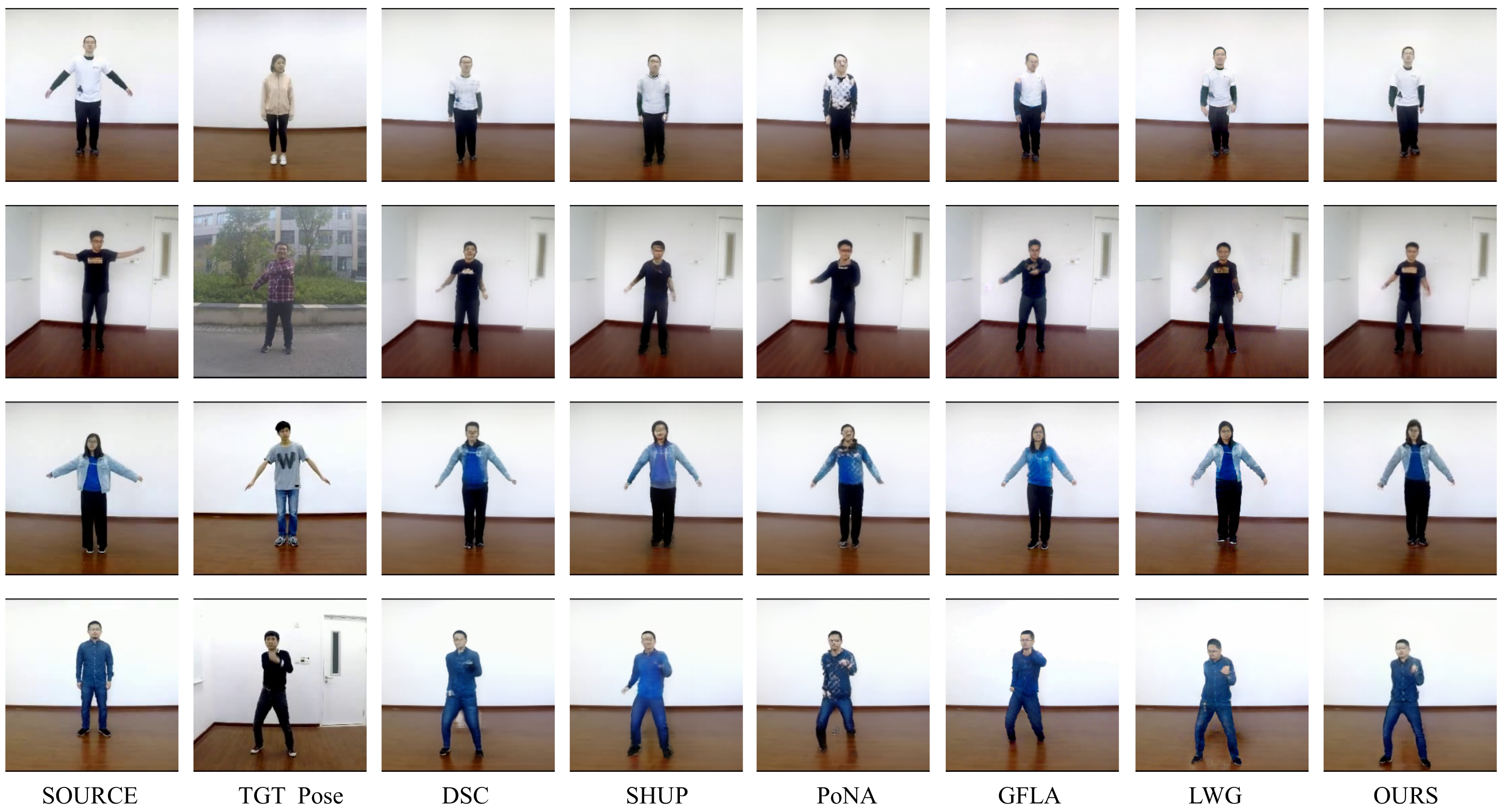}
\end{center}
\caption{Qualitative comparison of our proposed method against existing representative methods on \textit{iPER} dataset in cross-transfer setting. We find that results of our methods perceptually retain the most details and facial characteristics of source images among all methods, although there are no ground truths for comparison under this transfer setting.
}
\label{fig:vis-ci}
\end{figure*}

\begin{table*}[htbp]
  \centering
    \begin{tabular}{lccccccccc}
    \toprule
    \multirow{2}[4]{*}{} & \multicolumn{4}{c}{self-transfer} & \multicolumn{5}{c}{cross-transfer} \\
\cmidrule(lr){2-5}    \cmidrule(lr){6-10}  
Methods & SSIM $\uparrow$ & LPIPS $\downarrow$ & face-CS $\uparrow$ & OS-CS-reid $\uparrow$      & fid $\downarrow$ & OS-freid $\downarrow$ & OS-CS-reid $\uparrow$ & face-FD $\downarrow$ & face-CS $\uparrow$ \\
    \midrule
    PG2,\cite{ma2017pose}    & \underline{0.852} & 0.133 & 0.280  & 0.698  & 126.486 & 185.578 & 0.592 & 171.887 & 0.382 \\
    SHUP,\cite{balakrishnan2018synthesizing}   & 0.815 & 0.0961 & 0.312 & 0.751  & 75.753 & 125.576 & {0.611} & 127.472 & 0.365 \\
    DSC,\cite{siarohin2018deformable}   & 0.787 & 0.2370 & 0.293 & 0.669  & 156.967 & 205.820 & 0.590  & 204.565 & {0.407} \\
    
    PoNA, \cite{li2020pona}  & \textbf{0.858}  &  0.0900 & \underline{0.418} & \underline{0.813} & 81.876  & \underline{95.122}  & \underline{0.649}  & 32.216  & \underline{0.473} \\
     
    GFLA,\cite{ren2020deep}  & 0.833  & 0.0900  &  0.396  & 0.775 & 67.659  & 105.762  & 0.620  & \textbf{30.676} & 0.390 \\

    LWG,\cite{liu2019liquid}   & 0.820  & \textbf{0.0829} & {0.409} & {0.769}  & \underline{55.198} & {115.897} & 0.603 & {37.622} & 0.402 \\
    OURS      & {0.850}  & \underline{0.0859} & \textbf{0.442} & \textbf{0.828}  & \textbf{50.287} & \textbf{73.407} & \textbf{0.657} & \underline{32.037} & \textbf{0.497} \\
    \bottomrule
    \end{tabular}%
    \caption{Quantitative comparison of our proposed framework with previous state-of-the-art methods on the \textit{iPER} dataset for pose transfer in two transfer settings. The best performance is in \textbf{bold}, and the second best is \underline{underlined}. This table shows that for most metrics, our method achieves the best performance, and for the remaining metrics, our method ranks second. For all reid-based metrics, our method has best performance, which indicates that our method can effectively preserve pose-invariant visual appearance features of the source person thus it can facilitate the human pose transfer task as a reliable data augmentation technique.   
    }
  \label{tab:quancom}%
\end{table*}

\subsection{Summary}
Here, we summarize the proposed lifting-and-projection framework. This framework splits the task of pose transfer into foreground and background generation with four functional modules: mesh recovery, background generation, foreground generation and fusion modules. To achieve our goal of learning the foreground appearance feature in the mesh space, we design LPNet for the foreground generation module, in which foreground features are processed with lifting, 3D processing and projection operations.
In addition, we design ADCNet to enhance the output feature of LPNet. The loss functions for training the model are also discussed.

\section{Experimental Results}
\label{sec:experiments}

In this section, we conduct experiments to validate the proposed lifting-and-projection framework for human pose transfer. 
We compare our model with existing methods including mesh-based model and image-translation-based models.
In addition, we conduct an ablation study to bring more analysis on the efficacy of the core component of our proposed framework. The experimental results demonstrate the effectiveness of our proposed framework in both qualitative visual perception and quantitative metric values.

\subsection{Dataset}
We perform experiments on the Impersonator (iPER) dataset  \cite{liu2019liquid} and Fashion dataset \cite{zablotskaia2019dwnet}.
The iPER dataset contains videos of 30 subjects and 103 clothing items. 
iPER provides poses in the form of both body meshes and keypoints, which enables us to compare different types of human pose transfer models.  We use 256 $\times$ 256-resolution video for both training and testing.
For the training and validation split, we simply follow \cite{liu2019liquid}. 
The Fashion dataset contains 600 short videos, 500 videos for training and the remaining 100 videos for testing. Each video consists of approximately 350 frames.

\subsection{Experimental setup}

\textbf{Training details}. Our model is implemented in PyTorch. 
The weight hyperparameters $\lambda_{rec}$, $\lambda_{perc}$, $\lambda_{adv}$, $\lambda_{mask}$ and $\lambda_{face}$ are set to 10.0, 10.0, 1.0, 1.0 and 5.0, respectively.
Both the generator and discriminator are trained by the Adam optimizer for 30 epochs, where the training pairs are randomly sampled from each training video. The learning rate for the discriminator is kept $2\times 10^{-3}$ during training, while for the generator, it is initialized to $2\times 10^{-3}$ and kept for 5 epochs before linearly decaying to $2\times 10^{-6}$.

\textbf{Baseline methods}. We compare our proposed framework with several representative works on human pose transfer, including the mesh-based method LWG \cite{liu2019liquid}, and image-translation-based methods: PG2 \cite{ma2017pose} , SHUP \cite{balakrishnan2018synthesizing}, DSC \cite{siarohin2018deformable}, PONA \cite{li2020pona} and GFLA \cite{ren2020deep} .

We reproduce their model using the code they released and model weights if available and compare our model with them under the same protocol.

\textbf{Evaluation metrics}.
Although evaluating the image quality is a subjective process, 
in addition to evaluating the image quality subjectively, we adopt several commonly used quantitative metrics to compare different models. 
In our experiments, we consider two human pose transfer settings: self-transfer and cross-transfer.  Self-transfer means that the source image and the target pose come from the same person, while in the cross-transfer setting, the source image and the target pose come from different persons. 

For self transfer, we use structural similarity (SSIM) \cite{wang2004image} and learned perceptual image patch similarity (LPIPS) \cite{zhang2018unreasonable} to measure the similarity between the generated target image and the real image. 
SSIM is a classical and widely used method to measure the patch-wise structural similarity between two images.
LPIPS evaluates the distance between two image patches by comparing their deep features.
These two metrics can fully utilize the ground-truth target images that are available only in the self-transfer scenario. 
Generally, human visual system (HVS) has particular salient regions when assessing the image quality.
Therefore, to compare the performances on the generation details of human faces and human body parts between different models, we borrow the metrics face-OS and OS-CS-reid released by the iPER dataset~\cite{liu2019liquid}.
We utilize these two metrics to measure the feature dissimilarity of the cropped face region and person region separately between the synthesized image and the ground-truth image.      

For cross transfer, we adopt three metrics based on the Fréchet inception distance: fid, PCB-CS-reid and OS-freid. These metrics are based on comparing the deep features of cropped images obtained from neural networks, and their major differences lie in different choices of backbone network from different person-reidentification models. 
In addition, we use two metrics based on cosine similarity, namely, face-CS and OS-CS-reid \cite{liu2019liquid}, which measure the feature cosine similarity of the cropped face regions and cropped person regions between the source image and synthesized image.
Note that we do not employ the metric of inception score (IS), as 
it is somewhat unreliable \cite{barratt2018note} and irrelevant to the human pose transfer task.

\subsection{Comparison with existing methods on the iPER dataset}
\subsubsection{Qualitative Results}
We visualize some representative synthesized images generated by different methods to present a qualitative comparison.
Fig. \ref{fig:vis-si} and Fig. \ref{fig:vis-ci} show comparisons of the self-transfer and cross-transfer settings. 
In Fig. \ref{fig:vis-si}, the two left columns present the source images and target images (also ground-truth for self-transfer setting), and the other columns from left to right are generations of different methods. 
Fig. \ref{fig:vis-ci} shows the source images, target pose images and synthesized results of different methods in the cross-transfer setting and there are no ground-truth images in this setting.

In both transfer settings, the advantages of our method compared with other methods are as follows: 
(1) Appearance-keeping: The generated images preserve much information on the person appearance of the source image, such as the clothing style and shoe color, especially clothing details, such as small logos or patterns within clothes. 
(2) Identity-preserving: Facial details are faithful to source images, including hair, face shape, silhouette and accessories such as glasses. In addition, blurring effects in face are common among all methods, but less common in our results.
This effect is probably due to few pixels in the source face and the large pose variation. 
(3) Realistic-looking: Our synthesized images appear sharper, more harmonious and closer to natural images. 
Additionally, we observe that all of these methods perform worse in cross transfer setting than self transfer setting, which is reasonable, as cross-transfer is more challenging.     

We show some difficult transfer cases of the iPER dataset in Fig. \ref{fig:intro-results}. 
From an overall perspective, we observe that mesh-based methods, including LWG \cite{liu2019liquid} and our model, perform better than image-translation-based DSC \cite{siarohin2018deformable}, SHUP \cite{balakrishnan2018synthesizing}, PoNA \cite{li2020pona} and GFLA \cite{ren2020deep}.
This result shows the effectiveness of mesh-based methods in human pose transfer.
Through the comparison, we also find that our method is more capable than LWG for handling difficult cases.
As shown in the red box of the third row in Fig. \ref{fig:intro-results}, our model can clearly preserve and infer the clothing details without erasing it or expanding the area as LWG does.
This finding demonstrates the superiority of our method in generating details of target images. 

Since LWG and our model rely on sampling from the source image, this can cause some unrealness in the synthesized image in some cases, such as when there is occlusion in the source pose. 
This effect can be seen in the first and second rows of LWG generated results in Fig. \ref{fig:intro-results}.
However, this effect does not occur in the synthesized images of our method.
Our method can mitigate this problem mainly because we incorporate graph convolution networks and utilize information aggregation and diffusion in a region-wise and patch-wise manner to address the problems caused by occlusion and erroneous sampling.

\begin{figure*}[h]
\begin{center}
\includegraphics[width=0.9\linewidth]{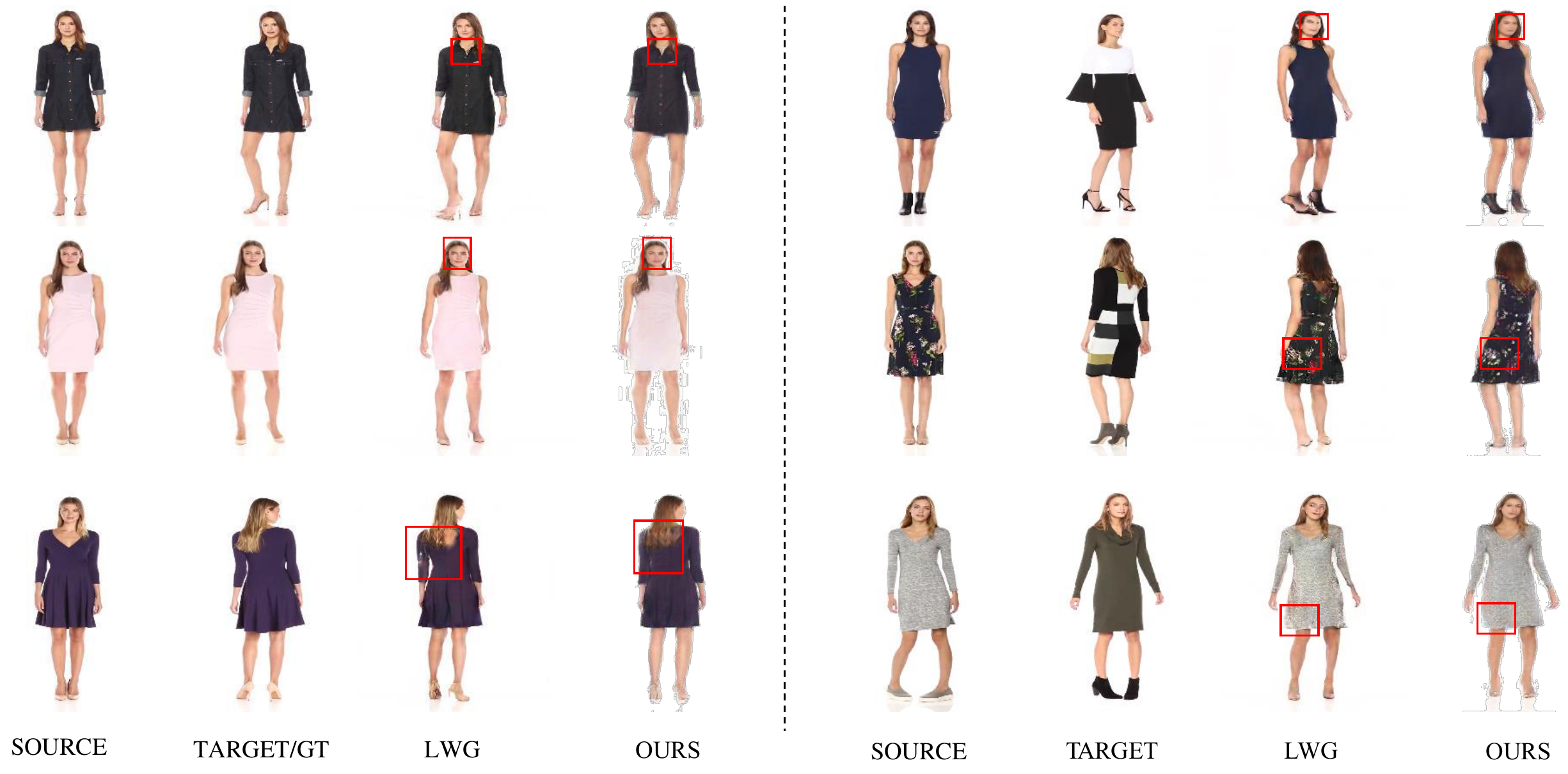}
\end{center}
\caption{Qualitative comparison between two mesh-based methods: liquid warping GAN (LWG) \cite{liu2019liquid} and our proposed method (OURS). Experiments are conducted on \textit{Fashion} dataset in two transfer settings: the left part of the illustration shows the self-transfer results and the right part shows the cross-transfer results.}
\label{fig:vis-fash-comp}
\end{figure*}

\subsubsection{Quantitative results}
In addition to qualitative analysis, we perform quantitative comparisons between different approaches including PG2\cite{ma2017pose}, DSC\cite{siarohin2018deformable}, SHUP\cite{balakrishnan2018synthesizing}, PoNA\cite{li2020pona}, GFLA\cite{ren2020deep}, LWG\cite{liu2019liquid} and our approach using the aforementioned metrics. 
We employ the same testing protocol to generate testing pairs.
Table \ref{tab:quancom} shows that our model achieves consistently better performance compared with existing methods including both image-based methods and mesh-based methods. 
Our model obtains the best or second best results in most metrics in both transfer settings. 
For the SSIM metric, LWG appears to acquire worse results but our model can still have a high SSIM score and be very close to the best results. This observation demonstrates that our proposed method preserves the structural information more effectively. 
For the reid-based metrics, our model acquires nearly all the best results.  
This finding indicates that our model holds the visual appearances nicely and preserves the identity before and after the transfer.
This advantage of our method demonstrates that our approach can be used as a reliable data augmentation technique for person reidentification task \cite{zhao2017deeply,deng2018image,zeng2020illumination,xie2020progressive}.

\begin{figure}[tb]
\begin{center}
\includegraphics[width=0.85\linewidth]{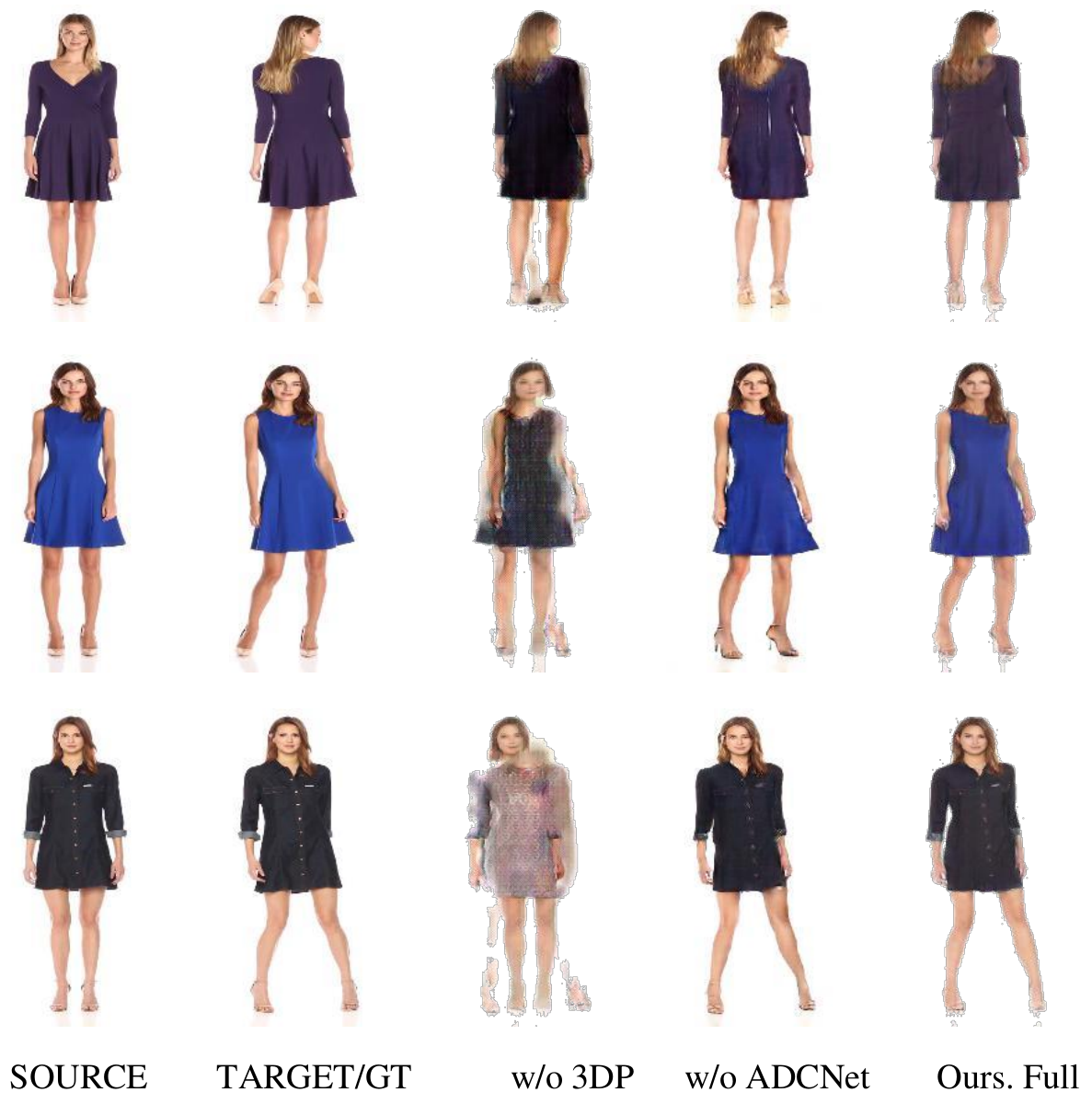}
\end{center}
\caption{Qualitative results of the ablation study on the \textit{Fashion} dataset. SOURCE column includes source images. The images belonging to the TARGET/GT column provide target poses for transfer and the images are also ground-truth in the self transfer setting. The three columns on the right showcase the transfer results of different model variants.}
\label{fig:ablation}
\end{figure}

\subsection{Comparison between two mesh-based methods on the Fashion dataset}
In addition, to have a deep comparison between two mesh-based methods, namely, LWG \cite{liu2019liquid} and our proposed method, we conduct further experiments on the Fashion dataset \cite{zablotskaia2019dwnet}.
We implement two mesh-based methods on the Fashion dataset and we adopt the same evaluation protocols and metrics as performed on the iPER dataset.

Table \ref{tab:mesh-method-com} lists quantitative results under different metrics of LWG and our proposed method in two transfer settings. For most of the metrics, the performance of our method surpasses that LWG. 
Fig. \ref{fig:vis-fash-comp} illustrates some qualitative results of two mesh-based methods. 
Although some pose variations are relatively large and the textures of source images are complex, both mesh-based methods can generate satisfactory images that conform to target poses.
However, compared with LWG, the results of our method show fewer artifacts.
The face parts of foreground persons synthesized by our method are more realistic than those synthesized by LWG.

\begin{table}[tb]
  \centering

\setlength{\tabcolsep}{7pt}{
    \begin{tabular}{lcccc}
    \toprule
    \multirow{2}[2]{*}{} & \multicolumn{4}{c}{self transfer} \\
\cmidrule{2-5}     
Methods & SSIM $\uparrow$ & LPIPS $\downarrow$ & face-CS $\uparrow$ & OS-CS-reid $\uparrow$      \\
    \midrule
    LWG \cite{liu2019liquid}    &   0.838  &  0.0952  & 0.396    &   0.817  \\
    OURS  &  \textbf{0.840}  & \textbf{0.0934} & \textbf{0.405} & \textbf{0.823}   \\

\bottomrule
    \end{tabular}%
}

\setlength{\tabcolsep}{2.5pt}{
\begin{tabular}{lccccc}
 \multirow{2}[2]{*}{}  & \multicolumn{5}{c}{cross transfer} \\
 \cmidrule{2-6}
 Methods & fid $\downarrow$ & OS-freid $\downarrow$ & OS-CS-reid $\uparrow$ & face-FD $\downarrow$ & face-CS $\uparrow$ \\
 \midrule
    LWG \cite{liu2019liquid}  & \textbf{30.895} & \textbf{84.312}  & 0.688 & 171.389  &  0.354 \\
    OURS     & 31.597  & 87.434   & \textbf{0.695}    & \textbf{150.537}   &  \textbf{0.364}         \\
  
\bottomrule
 
\end{tabular}%
} 
    \caption{Quantitative comparison between two mesh-based methods: LWG \cite{liu2019liquid} and our proposed LPNet (OURS) on \textit{Fashion dataset}. The best performance for each metric is in \textbf{bold}. 
    }
  \label{tab:mesh-method-com}%
\end{table}

\subsection{Ablation study}
\label{sec:ablation}
Here we perform an ablation study to verify the effectiveness of different components in our proposed method.
We design two baseline models, namely, a) \textit{LPNet w/o. 3DP} and b) \textit{LPNet only}, and make a comparison with our full model, \textit{LPNet $+$ ADCNet}. The details are described below.

\textbf{LPNet w/o. 3DP.} In this baseline, under the lifting-and-projection framework, we implement a foreground generator with only LPNet variants without 3D processing, where only lifting and projection operations are performed in the LPBs. In this model, neither structural nor spatial information of the human mesh is captured, LPNet without 3D processing only performs spatial transforms like DSC \cite{siarohin2018deformable}.

\textbf{ LPNet only.} In this model, we remove the ADCNet branch in foreground generator to investigate its effects on the synthesized images.

\textbf{LPNet$+$ADCNet (Full).} Our proposed full model utilizes GCNs to process and learn features in the 3D mesh space, and uses the ADCNet branch to compensate the texture details into target image images directly from the source image, as mentioned in Section \ref{sec:networks}. 

The training strategy and loss functions for all these models are the same. Fig.\ref{fig:ablation} and Table \ref{tab:ablation} show the qualitative and quantitative results for this ablation study. By comparing the results of the three models, we observe that our full model outperform the other variants both quantitatively and qualitatively.
Three-dimensional processing with graph convolutions and ADCNet can enhance the results.

For the baseline model \textit{LPNet w/o 3DP}, 
the poor learning ability of the foreground generator leads the model to rely heavily on the background generator, producing underfitted results.

Compared with the \textit{LPNet only} model, our full model can synthesize images with more authentic detailed textures that are more consistent with source images with fewer artifacts on body regions.
This finding demonstrates the effectiveness of the ADCNet branch in our model design.
The ablation study supports the effectiveness of the components of our proposed method.

\begin{table}[tb]
  \centering
\setlength{\tabcolsep}{7pt}{
    \begin{tabular}{lcccc}
    \toprule
    \multirow{2}[2]{*}{} & \multicolumn{4}{c}{Self transfer} \\
\cmidrule{2-5}   
Methods & SSIM $\uparrow$ & LPIPS $\downarrow$ & face-CS $\uparrow$ & OS-CS-reid $\uparrow$\\
    \midrule
    LPNet w/o. 3DP    &  0.752  &  0.214  & 0.206    &   0.671  \\
    LPNet only   &  0.833   &  0.104 & 0.392   &  0.792         \\
    LPNet$+$ADCNet     & \textbf{0.840} & \textbf{0.0934} & \textbf{0.405}  & \textbf{0.823}   \\ 
\bottomrule
    \end{tabular}%
}

\setlength{\tabcolsep}{2.5pt}{
\begin{tabular}{lccccc}
 \multirow{2}[2]{*}{}  & \multicolumn{5}{c}{Cross transfer} \\
 \cmidrule{2-6}  
Methods  & fid $\downarrow$ & OS-freid $\downarrow$ & OS-CS-reid $\uparrow$ & face-FD $\downarrow$ & face-CS $\uparrow$ \\
 \midrule
    LPNet w/o. 3DP & 79.134 & 178.368  & 0.643 & 218.046  &  0.320 \\

    LPNet only    &  33.946   & 112.134  &  0.669  &  158.580  & 0.336              \\
    LPNet$+$ADCNet   & \textbf{31.597} & \textbf{87.434} & \textbf{0.695}  & \textbf{150.537}   & \textbf{0.364}    \\
    
\bottomrule
 
\end{tabular}%
} 
    \caption{Quantitative results of the ablation study of our proposed method on \textit{Fashion} dataset. The best performance is in \textbf{bold}. 
    }
  \label{tab:ablation}%
\end{table}

\subsection{Other issues} \label{sec:fail-case}
\subsubsection{Failure cases}

\begin{figure}
\centering    
\subfigure[Example 1: the synthesized person has an incorrect body orientation. This is caused by mesh estimator, which predicts an incorrect pose orientation from the target pose image. 
]{\label{fig:fail-a}\includegraphics[width=70mm]{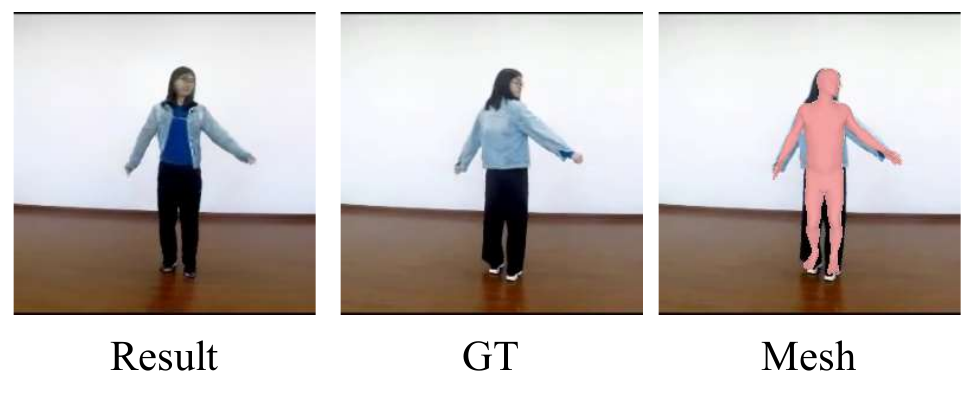}}
\subfigure[Example 2: the left hand is lost in the synthesized image of the first row. This is attributed to the intersection of the hand and body of the reconstructed mesh. In the second row we show our solution: by manually adjusting the pose parameter $\theta_t$, our model produces a reasonable result.  ]{\label{fig:fail-b}\includegraphics[width=70mm]{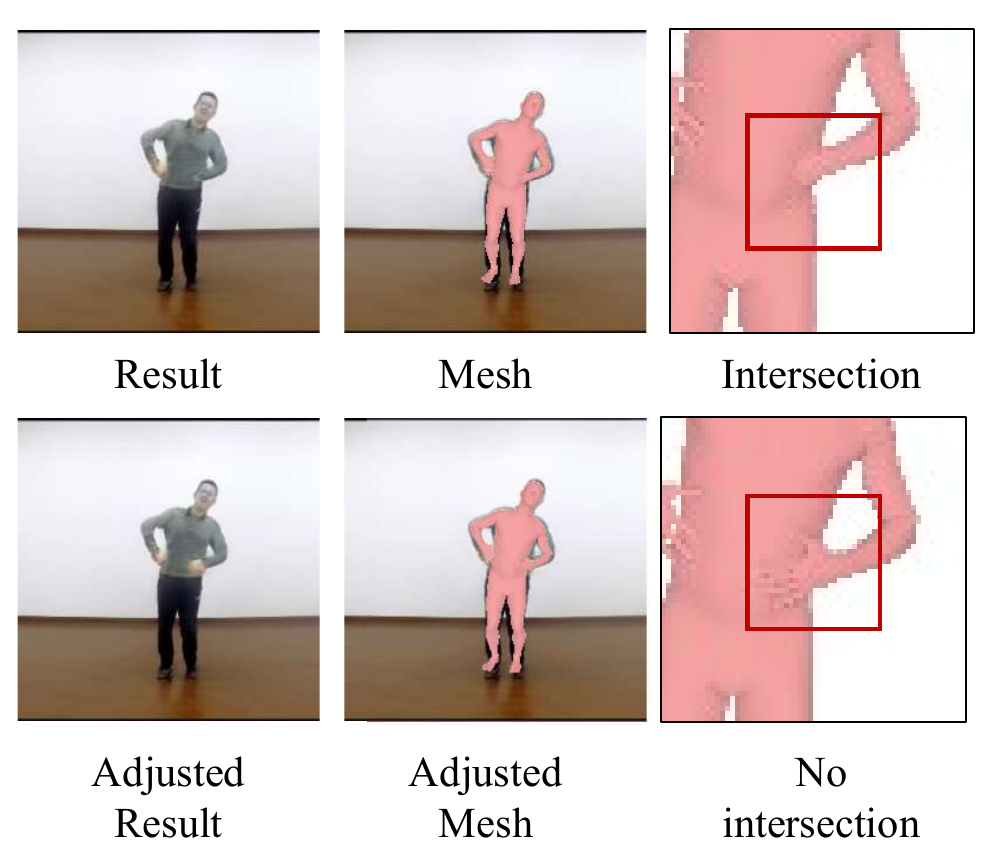}}
\caption{ Failure case study: two failure cases we found in our experimental results on the \textit{iPER} dataset. (a) Example 1 is a failure caused by the incorrect mesh estimated result. (b) Example 2 is a mesh intersection problem. }
\label{fig:fail-case}
\end{figure}

Although the results of our method are impressive, we still observe two failure cases.
We demonstrate the failure cases of our results in Fig. \ref{fig:fail-case}.

In Example 1 shown in Fig. \ref{fig:fail-a}, 
the target image is the person whose back is facing the camera.
However, the person in the synthesized image faces the camera.
This issue is caused by the wrong pose estimation result as shown in the third  subfigure of Fig. \ref{fig:fail-a}.

Example 2 is shown in Fig. \ref{fig:fail-b}. In the first row of Fig. \ref{fig:fail-b}, MRM produces a mesh whose body is intersected by the left hand in the cross transfer setting. As a result, the left hand is lost in the synthesized image. 
The intersection problem is caused by a simple combination of $\beta_s$ and $\theta_t$ from different persons in MRM. 
In practice, the intersection problem can be solved by adjusting the pose parameters $\theta_t$ with human interaction. 
We visualize the adjusted mesh and resynthesized image in the second row of Fig. \ref{fig:fail-b}. 
The results demonstrate that our model can produce a more reasonable and satisfactory result with a refined nonintersecting mesh.

\subsubsection{Running time}\label{sec:run}

We test the running time of our model on the platform with: an Intel(R) Xeon(R) Silver 4110 CPU $@$ 2.10 GHz and one Nvidia 1080Ti GPU.
The testing sample images are from the iPER dataset with a resolution of $256\times 256$. 

From Table \ref{tab:time-module}, we can see that although our core foreground generation module takes most of the computation overhead, and mesh recovery module (MRM) without a learning process also involves much processing time. This can be further improved.
 
From Table \ref{tab:time-fore}, we learn that graph convolutions (GCN) require a large computational overhead.
In addition, the renderer for projection takes up nearly 1/3 of the total time for foreground generation.  

\begin{table}[!ht]

    \centering
    \begin{tabular}{cr}
      \toprule
      Module & Time (ms) \\
      \midrule
     MRM  & 46.98 \\
     
     Background Generator & 11.68  \\
     Foreground Generator & 99.67  \\
     \midrule
         Total & 158.33  \\
      \bottomrule
      \end{tabular}
    \caption{Module-wise computation time of our model.}
      \label{tab:time-module}
\end{table}

\begin{table}[!ht]
    \centering
    \begin{tabular}{lr}
      \toprule
        Item & Time (ms) \\
      \midrule
    Encoder & 1.94  \\
    Lifting & 3.99   \\
    GCN & 47.99    \\
    Projection & 31.96   \\
    Decoder & 8.70      \\
    ADCNet & 4.81     \\
    \midrule
    Foreground Generator total & 99.67   \\
      \bottomrule
      \end{tabular}
    \caption{Item-wise computation time of the foreground generation module.}
      \label{tab:time-fore}
\end{table}%

\section{Conclusions}
\label{sec:conclusions}

In this work, we have proposed a lifting-and-projection framework for human pose transfer.
The core of our proposed LPNet for foreground generation consists of cascaded lifting-and-projection blocks (LPBs). 
Each block goes through a process of lifting source image features to mesh space, processing in 3D mesh space and reprojecting to 2D plane. 
This lifting-and-projection pipeline fully exploits the structural information of the reconstructed human body mesh in 3D space to obtain a comprehensive and generalized feature representation for the target pose image. Additionally, considering the natural graph structure and pose-invariant properties of the human body mesh, we deploy graph convolutional networks to process visual features attached to each node defined by the mesh vertex. 
Through information diffusion and aggregation of graphs, an enhanced and contextual feature representation for the target image is obtained with the guidance of a reconstructed human body mesh.
In addition, we design another network, called ADCNet, as an auxiliary branch for LPNet to compensate for the texture and appearance by modulating its output feature with the source foreground image. The experimental results demonstrate the superiority of our lifting-and-projection framework quantitatively and perceptually in two transfer settings on the iPER and Fashion datasets, compared with previous methods. 

For future work, the current performance of our method can be improved from two aspects.
1) Design a more delicate network branch from 2D level to preserve appearance details from source image as much as possible.
2) Eliminate the failure case introduced by mesh intersection problem with more advanced mesh editing methods. 
Furthermore, we believe that the idea behind our lifting-and-projection framework can provide new insights and be applied to related tasks such as garment transfer and face animation.

\bibliographystyle{IEEEbib}
\bibliography{ref}

\end{document}